\documentclass[runningheads]{llncs}

\frontmatter  \mainmatter

\usepackage[dvips]{graphicx}
\usepackage{epsfig}
\usepackage{here,wrapfig}

\input{psfig.sty}

\usepackage{epsbox,epsf}
\usepackage{amsmath,amssymb,bm} 

\usepackage{tabularx}
\usepackage{multirow}
\usepackage{colortbl}
\usepackage{algorithm}
\usepackage{algorithmic}
\usepackage{subfigure}
\usepackage[table]{xcolor}

\usepackage{color}
\usepackage[normalem]{ulem}
\definecolor{cyan}{cmyk}{1,0,0,0}

\begin{document}

\title{An experimental comparison of hybrid \\
  algorithms for Bayesian network \\
  structure learning}
  
\titlerunning{Comparison of hybrid algorithms for Bayesian network structure learning}

\author{
  Maxime Gasse
  \and
  Alex Aussem
  \and
  Haytham Elghazel
}

\authorrunning{Maxime Gasse}

\institute{
Universit\'e de Lyon, CNRS\\
Universit\'e Lyon 1, LIRIS, UMR5205, F-69622, France\\
}

\maketitle

\begin{abstract}
We present a novel hybrid algorithm for Bayesian network structure
learning, called Hybrid HPC (H2PC). It first reconstructs the skeleton
of a Bayesian network and then performs a Bayesian-scoring greedy
hill-climbing search to orient the edges. It is based on a subroutine
called HPC, that combines ideas from incremental and divide-and-conquer
constraint-based methods to learn the parents and children of a target
variable.  We conduct an experimental comparison of H2PC against
Max-Min Hill-Climbing (MMHC), which is currently the most powerful
state-of-the-art algorithm for Bayesian network structure learning, on
several benchmarks with various data sizes. Our extensive experiments
show that H2PC outperforms MMHC both in terms of goodness of fit to
new data and in terms of the quality of the network structure itself,
which is closer to the true dependence structure of the data. 
The source code (in \textit{R}) of H2PC as well as all data sets used for the empirical
tests are publicly available.
\end{abstract}

\section{Introduction}

A Bayesian network (BN) is a probabilistic model formed by a structure
and parameters. The structure of a BN is a directed acyclic graph
(DAG), whilst its parameters are conditional probability distributions
associated with the variables in the model. The graph of a BN itself
is an independence map, which is very useful for many applications,
including feature selection \cite{Aliferis10a,Pen07,Morais10a}
and inferring causal relationships from observational data
\cite{Byron08,Aliferis10a,Aussem12a,Aussem10c,Cawley08,Brown08}. The
problem of finding the DAG that encodes the conditional independencies
present in the data attracted a great deal of interest over the last
years \cite{Morais10b,Scutari10,Scutari12,Kojima10,Perrier08,Villanueva12,Pen12}.

Ideally the DAG should coincide with the dependence structure of the
global distribution, or it should at least identify a distribution as
close as possible to the correct one in the probability space. This
step, called structure learning, is similar in approaches and
terminology to model selection procedures for classical statistical
models. Basically, constraint-based (CB) learning methods systematically
check the data for conditional independence relationships and use them
as constraints to construct a partially oriented graph representative
of a BN equivalence class, whilst search-and-score (SS) methods make
use of a goodness-of-fit score function for evaluating graphical
structures with regard to the data set. Hybrid methods attempt to get
the best of both worlds: they learn a skeleton with a CB approach and
constrain on the DAGs considered during the SS phase. There are many
excellent treatments of BNs which survey the learning methods (see
\cite{Koller09} for instance). 

Both CB and SS approaches have advantages and disadvantages. CB
approaches are relatively quick, deterministic, and have a well defined
stopping criterion; however, they rely on an arbitrary significance
level to test for independence, and they can be unstable in the sense
that an error early on in the search can have a cascading effect that
causes many errors to be present in the final graph. SS approaches
have the advantage of being able to flexibly incorporate users'
background knowledge in the form of prior probabilities over the
structures and are also capable of dealing with incomplete records in
the database (e.g. EM technique). Although SS methods are favored in
practice when dealing with small dimensional data sets, they are slow
to converge and the computational complexity often prevents us from
finding optimal BN structures \cite{Perrier08}. With currently
available exact algorithms \cite{Koivisto04,Silander06} and a
decomposable score like BDeu, the computational complexity remains
exponential, and therefore, such algorithms are intractable for BNs
with more than around 30 vertices on current workstations
\cite{Kojima10}. For larger sets of variables, the computational
burden becomes prohibitive. With this in mind, the ability to restrict
the search locally around the target variable is a key advantage of CB
methods over SS methods. They are able to construct a local graph
around the target node without having to construct the whole BN first,
hence their scalability \cite{Pen07,Morais10a,Morais10b,Tsamardinos06,Pena08}. 

With a view to balancing the computation cost with the desired accuracy
of the estimates, several hybrid methods have been proposed recently.
Tsamardinos et al. \cite{Tsamardinos06} proposed the Min-Max Hill Climbing (MMHC)
algorithm and conducted one of the most extensive empirical comparison
performed in recent years showing that MMHC was the fastest and the
most accurate method in terms of structural error based on the
structural hamming distance. More specifically, MMHC outperformed both
in terms of time efficiency and quality of reconstruction the PC
\cite{Spi00}, the Sparse Candidate \cite{Friedman99}, the Three Phase
Dependency Analysis \cite{Cheng02}, the Optimal Reinsertion
\cite{Moore03}, the Greedy Equivalence Search \cite{Chickering02},
and the Greedy Hill-Climbing Search on a variety of networks, sample
sizes, and parameter values. Although MMHC is rather heuristic by
nature (it returns a local optimum of the score function), MMHC is
currently considered as the most powerful state-of-the-art algorithm
for BN structure learning capable of dealing with thousands of nodes
in reasonable time. With a view to enhance its performance on small
dimensional data sets, Perrier et al. \cite{Perrier08} proposed recently a hybrid
algorithm that can learn an \textit{optimal} BN (i.e., it converges
to the true model in the sample limit) when an undirected graph is
given as a structural constraint. They defined this undirected graph
as a super-structure (i.e., every DAG considered in the SS phase is
compelled to be a subgraph of the super-structure). This algorithm can
learn optimal BNs containing up to 50 vertices when the average degree
of the super-structure is around two, that is, a sparse structural
constraint is assumed. To extend its feasibility to BN with a few
hundred of vertices and an average degree up to four, \cite{Kojima10}
proposed to divide the super-structure into several clusters and
perform an optimal search on each of them in order to scale up to
larger networks. Despite interesting improvements in terms of score
and structural hamming distance on several benchmark BNs, they report
running times about $10^3$ times longer than MMHC on average, which
is still prohibitive in our view. 

Therefore, there is great deal of interest in hybrid methods capable of
improving the structural accuracy of both CB and SS methods on graphs
containing up to thousands of vertices. However, they make the strong
assumption that the skeleton (also called super-structure) contains at
least the edges of the true network and as small as possible extra
edges. While controlling the false discovery rate (i.e., false extra
edges) in BN learning has attracted some attention recently
\cite{ArmenT11,Pena08,Tsamardinos08}, to our knowledge, there is no
work on controlling actively the rate of false-negative errors (i.e.,
false missing edges). 

In this study, we compare MMHC with a another hybrid algorithm for BN
structure learning, called Hybrid HPC (H2PC). H2PC and MMHC share
exactly the same SS procedure to allow for fair comparisons; the only
difference lies in the procedure for learning the skeleton (i.e., the
undirected graph given as a structural constraint to the SS search).
While MMHC is based on Max-Min Parents and Children (MMPC) to learn the
parents and children of a variable, H2PC is based on a subroutine
called Hybrid Parents and Children (HPC), that combines ideas from
incremental and divide-and-conquer CB methods.  The ability of HPC and
MMPC \cite{Tsamardinos06} to infer the parents and children set of
candidate nodes was assessed in \cite{Morais10b} through several
empirical experiments. In this work, we conduct an experimental
comparison of H2PC against Max-Min Hill-Climbing (MMHC) on several
benchmarks and various data sizes.

\section{Preliminaries}

Formally, a BN is a tuple $<\mathbb{G}, P>$, where $\mathbb{G} =
<\mathbf U, \mathbf E>$ is a directed acyclic graph (DAG) whose nodes
represent the variables in the domain $\mathbf{U}$, and whose edges
represent direct probabilistic dependencies between them. $P$ denotes
the joint probability distribution on $\mathbf{U}$. The BN structure
encodes a set of conditional independence assumptions: that each node
$X_i$ is conditionally independent of all of its non descendants in
$\mathbb{G}$ given its parents $\mathbf{Pa}_i^\mathbb{G}$. These
independence assumptions, in turn, imply many other conditional
independence statements, which can be extracted from the network using
a simple graphical criterion called d-separation \cite{Pea88}. 

We denote by $X \perp_P Y|\mathbf Z$ the conditional independence
between $X$ and $Y$ given the set of variables $\mathbf Z$ where $P$
is the underlying probability distribution. Note that an exhaustive
search of \textbf{Z} such that $X \perp_P Y|\mathbf Z$ is a
combinatorial problem and can be intractable for high dimension data
sets. We use $X \perp_\mathbb{G} Y|\mathbf Z$ to denote the assertion
that $X$ is d-separated from $Y$ given $\mathbf Z$ in $\mathbb{G}$. We
denote by $\textbf{dSep}(X,Y)$, a set  that d-separates $X$ from $Y$.
If $<\mathbb{G}, P>$ is a BN,  $X \perp_P Y|\mathbf Z$ if $X
\perp_\mathbb{G} Y|\mathbf Z$. The converse does not necessarily hold.
We say that $<\mathbb{G}, P>$ satisfies the \emph{faithfulness
condition} if the d-separations in $\mathbb{G}$ identify \emph{all and
only} the conditional independencies in $P$, i.e., $X \perp_P
Y|\mathbf Z$ if and only if $X \perp_\mathbb{G} Y|\mathbf Z$. We denote
by $\textbf{PC}_X^{\cal G}$, the set of parents and children of $X$ in
${\cal G}$, and by $\textbf{SP}_X^{\cal G}$, the set of spouses of $X$
in ${\cal G}$, i.e., the variables that have common children with $X$.
These sets are unique for all ${\cal G}$, such that $<{\cal G},P>$
satisfies the faithfulness condition and so we will drop the
superscript ${\cal G}$.

\section{Constraint-based structure learning}

The induction of local or global BN structures is handled by CB methods
through the identification of local neighborhoods (i.e.,
$\textbf{PC}_X$), hence their scalability to very high dimensional
data sets. CB methods systematically check the data for conditional
independence relationships in order to infer a target's neighborhood.
Typically, the algorithms run either a $G^2$ or a $\chi^2$ independence
test when the data set is discrete and a Fisher's Z test when it is
continuous in order to decide on dependence or independence, that is,
upon the rejection or acceptance of the null hypothesis of conditional
independence. Since we are limiting ourselves to discrete data, both
the global and the local distributions are assumed to be multinomial,
and the latter are represented as conditional probability tables.
Conditional independence tests and network scores for discrete data are
functions of these conditional probability tables through the observed
frequencies $\{n_{ijk};i = 1,\ldots ,R;j = 1,\ldots, C;k = 1,\ldots, L\}$
for the random variables $X$ and $Y$ and all the configurations of the
levels of the conditioning variables $\mathbf{Z}$. We use $n_{i+k}$ 
as shorthand for the marginal $\sum_j n_{ijk}$ and similarly for $n_{i+k}, n_{++k}$ and $n_{+++}=n$.
We use a classic conditional independence test based on the mutual information. The
mutual information is an information-theoretic distance measure defined
as

$$
MI(X, Y |\mathbf{Z}) = \sum_{i=1}^R \sum_{j=1}^C \sum_{k=1}^L 
\frac{n_{ijk}}{n} \log{\frac{n_{ijk}n_{++k}}{n_{i+k}n_{+jk}}}
$$

It is proportional to the log-likelihood ratio test $G^2$ (they differ
by a 2n factor, where n is the sample size). The asymptotic null
distribution is $\chi^2$ with $(R-1)(C -1)L$ degrees of freedom. For a
detailed analysis of their properties we refer the reader to
\cite{Agresti02}. The main limitation of this test is the rate of
convergence to its limiting distribution, which is particularly
problematic when dealing with small samples and sparse contingency
tables. The decision of accepting or rejecting the null hypothesis
depends implicitly upon the degree of freedom which increases
exponentially with the number of variables in the conditional set.
Several heuristic solutions have emerged in the literature
\cite{Spi00,Morais10b,Tsamardinos06,Tsamardinos10} to overcome some
shortcomings  of the asymptotic tests. In this study we use the two
following heuristics that are used in MMHC.  First, we do not perform
$MI(X, Y |\mathbf{Z}) $ and assume independence if there are not enough
samples to achieve large enough power. We require that the average
sample per count is above a user defined parameter, equal to 5, as in
\cite{Tsamardinos06}. This heuristic is called the power rule. Second,
we consider as structural zero either case $n_{+jk}$ or $n_{i+k} = 0$. For
example, if $n_{+jk} = 0$, we consider y as a structurally forbidden value
for Y when Z = z and we reduce R by 1 (as if we had one column less in
the contingency table where Z = z).  This is known as the degrees of
freedom adjustment heuristic.

\section{The Hybrid Parents and Children algorithm (HPC)}

In this section, we present a brief overview of HPC. For further
details, the reader is directed to \cite{Morais10a,Morais10b} as well
as references therein. HPC (Algorithm \ref{HPC}) can be viewed as an
ensemble method for combining many weak PC learners in an attempt to
produce a stronger PC learner. HPC  is based on three subroutines:
\emph{Data-Efficient Parents and Children Superset} (DE-PCS),
\emph{Data-Efficient Spouses Superset} (DE-SPS), and \emph{Interleaved
Incremental Association Parents and Children} (Inter-IAPC), a weak PC
learner based on Inter-IAMB \cite{Tsamardinos03} that requires little
computation.  HPC may be thought of as a way to compensate for the
large number of false negatives, at the output of the weak PC learner,
by performing extra computations. It receives a target node \emph{T},
a data set ${\cal D}$ and a set of variables \textbf{U} as input and
returns an estimation of $\textbf{PC}_T$. It is hybrid in that it
combines the benefits of incremental and divide-and-conquer methods.
The procedure starts by extracting a superset $\textbf{PCS}_T$ of
$\textbf{PC}_T$ (line 1) and a superset $\textbf{SPS}_T$ of
$\textbf{SP}_T$ (line 2) with a severe restriction on the maximum
conditioning size ($\textbf{Z}<= 2$) in order to significantly increase
the reliability of the tests. A first candidate PC set is then obtained
by running the weak PC learner on $\textbf{PCS}_T \cup \textbf{SPS}_T$
(line 3). The key idea is the decentralized search at lines 4-8 that
includes, in the candidate PC set, all variables in the superset
$\textbf{PCS}_T \setminus \textbf{PC}_T$ that have $T$ in their vicinity.
Note that, in theory, $X$ is in the output of Inter-IAPC($Y$) if and
only if $Y$ is in the output of Inter-IAPC($X$). However, in practice,
this may not always be true, particularly when working in
high-dimensional domains. By loosening the criteria by which two nodes
are said adjacent, the effective restrictions on the size of the
neighborhood are now far less severe. The decentralized search has
significant impact on the accuracy of HPC. It enables the algorithm to
handle large neighborhoods while still being correct under faithfulness
condition. 

\begin{algorithm}
\begin{small}
\caption{\emph{HPC}}
\label{HPC}
\begin{algorithmic}[1]

           \REQUIRE
           $T$: target; ${\cal D}$: data set; \textbf{U}: the set of variables\\
           \ENSURE
           $\textbf{PC}_T$: Parents and Children of $T$\\
           \emph{}\\

           \STATE $[\textbf{PCS}_T,\textbf{dSep}] \leftarrow \emph{DE-PCS}(T,{\cal D})$
           \STATE $\textbf{SPS}_T \leftarrow \emph{DE-SPS}(T,{\cal D},\textbf{PCS}_T,\textbf{dSep})$

           \STATE $\textbf{PC}_T \leftarrow \emph{Inter-IAPC}(T,{\cal D},(T\cup\textbf{PCS}_T\cup \textbf{SPS}_T))$\\
           \FORALL {$X \in \textbf{PCS}_T\setminus\textbf{PC}_T$}
           \IF {$T\in\emph{Inter-IAPC}(X,{\cal D},(T\cup \textbf{PCS}_T\cup \textbf{SPS}_T))$}
           \STATE $\textbf{PC}_T \leftarrow \textbf{PC}_T \cup{X}$
           \ENDIF
           \ENDFOR

\end{algorithmic}
\end{small}
\end{algorithm}

\begin{algorithm}
\begin{small}
\caption{\emph{Inter-IAPC}}
\label{Inter-IAPC}
         \begin{algorithmic}[1]

           \REQUIRE
           $T$: target; \emph{D}: data set; $\textbf{U}$: set of variables; \\
           \ENSURE
           $\textbf{PC}_T$: Parents and children of $T$;\\
           \emph{}\\

           \STATE $\textbf{MB}_T \leftarrow \emptyset$\\

           \REPEAT
           \STATE \emph{\textbf{*} Add true positives to} $\textbf{MB}_T$\\
           \STATE $Y \leftarrow \textbf{argmax}_{\emph{X} \in {(\textbf{U} \setminus \textbf{MB}_T \setminus
           \emph{T})}}\emph{dep}(T,X|\textbf{MB}_T)$\\
           \IF {$T \not\perp Y | \textbf{MB}_T$}
           \STATE $\textbf{MB}_T \leftarrow \textbf{MB}_T \cup Y$\\
           \ENDIF

           \emph{ }\\
           \emph{\textbf{*} Remove false positives from} $\textbf{MB}_T$\\
           \FORALL {$\emph{X} \in \textbf{MB}_T$}
           \IF {$T\perp X|(\textbf{MB}_T\setminus{X})$}
           \STATE $\textbf{MB}_T \leftarrow \textbf{MB}_T \setminus \emph{X}$\\
           \ENDIF
           \ENDFOR
          \UNTIL {$\textbf{MB}_T$ has not changed}\\

          \emph{ }\\
          \emph{\textbf{*} Remove spouses of $T$ from} $\textbf{MB}_T$\\
          \STATE $\textbf{PC}_T \leftarrow \textbf{MB}_T$
          \FORALL {$X \in \textbf{MB}_T$}
          \IF {$\exists \textbf{Z} \subseteq (\textbf{MB}_T \setminus X)$ such that $T \perp X \mid \textbf{Z}$}
          \STATE $\textbf{PC}_T \leftarrow \textbf{PC}_T\setminus X$
          \ENDIF
          \ENDFOR

\end{algorithmic}
\end{small}
\end{algorithm}

\begin{algorithm}
\begin{small}
\caption{\emph{Hybrid HPC}}
\label{H2PC}
\begin{algorithmic}[1]

           \REQUIRE
           ${\cal D}$: data set; \textbf{U}: the set of variables\\
           \ENSURE
           A DAG ${\cal G}$ on the variables \textbf{U} \\
           \emph{}\\

           \FORALL {pair of nodes $X,Y \in \textbf{U}$} 
           \STATE Add X in $\textbf{PC}_Y$ and Add Y in $\textbf{PC}_X$ if $X \in HPC(Y)$ and $Y \in HPC(X)$
           \ENDFOR

          \STATE {Starting from an empty graph, perform greedy hill-climbing with operators \textit{add-edge, delete-edge, reverse-edge}.
          Only try operator \textit{add-edge} $X \rightarrow Y$ if $Y \in \textbf{PC}_X$}

\end{algorithmic}
\end{small}
\end{algorithm}

Inter-IAPC is a fast incremental method that receives a data set ${\cal
D}$ and a target node $T$ as its input and promptly returns a rough
estimation of $\textbf{PC}_T$, hence the term ``weak'' PC learner. The
subroutines DE-PCS and DE-SPS (omitted for brevity) search a superset
of $\textbf{PC}_T$ and $\textbf{SP}_T$ respectively with a severe
restriction on the maximum conditioning size ($|\textbf{Z}|<= 1$ in
DE-PCS and $|\textbf{Z}|<= 3$ in DE-SPS) in order to significantly
increase the reliability of the tests. The variable filtering has two
advantages : i) it allows HPC to scale to hundreds of thousands of
variables by restricting the search to a subset of relevant variables,
and ii) it eliminates many (almost) deterministic relationships that
produce many false negative errors in the output of the algorithm.
Again, the reader is encouraged to consult the papers by
\cite{Morais10a,Morais10b} for gaining more insight on these procedures.

\section{Hybrid HPC (H2PC)}

In this section, we discuss the SS phase. The following discussion
draws strongly on \cite{Tsamardinos06} as the SS phase in Hybrid HPC
and MMHC are exactly the same. The idea of constraining the search to
improve time-efficiency first appeared in the Sparse Candidate
algorithm \cite{Friedman99}. It results in efficiency improvements over
the (unconstrained) greedy search. All recent hybrid algorithms build
on this idea, but employ a sound algorithm for identifying the
candidate parent sets. The Hybrid HPC first identifies the parents and
children set of each variable, then performs a greedy hill-climbing
search in the space of BN. The search begins with an empty graph. The
edge addition, deletion, or direction reversal that leads to the
largest increase in score (the BDeu score was used) is taken and the
search continues in a similar fashion recursively. The important
difference from standard greedy search is that the search is
constrained to only consider adding an edge if it was discovered by HPC
in the first phase. We extend the greedy search with a TABU list
\cite{Friedman99}. The list keeps the last 100 structures explored.
Instead of applying the best local change, the best local change that
results in a structure not on the list is performed in an attempt to
escape local maxima. When 15 changes occur without an increase in the
maximum score ever encountered during search, the algorithm terminates.
The overall best scoring structure is then returned. Clearly, the more
false positives the heuristic allows to enter candidate PC set, the
more computational burden is imposed in the SS phase.

\section{Experimental validation}

\begin{table}
  \caption{Description of the BN benchmarks used in the experiments.}
  \label{tab:datasets}
  \begin{tabularx}{\textwidth}{XXXXXX}
  \hline
  \multicolumn{1}{c}{\multirow{2}{*}{network}} & \multicolumn{1}{c}{\multirow{2}{*}{\# of vars}} & \multicolumn{1}{c}{\multirow{2}{*}{\# of edges}} & \multicolumn{1}{c}{max. degree} & \multicolumn{1}{c}{\multirow{2}{*}{domain range}} & \multicolumn{1}{c}{min/med/max} \\
                                              &                                                  &                                                  & \multicolumn{1}{c}{in/out}      &                                                   & \multicolumn{1}{c}{PC set size} \\
  \hline
  \multicolumn{1}{c}{child}      & \multicolumn{1}{c}{20}   & \multicolumn{1}{c}{25}   & \multicolumn{1}{c}{2/7}  & \multicolumn{1}{c}{2-6}   & \multicolumn{1}{c}{1/2/8} \\
  \multicolumn{1}{c}{insurance}  & \multicolumn{1}{c}{27}   & \multicolumn{1}{c}{52}   & \multicolumn{1}{c}{3/7}  & \multicolumn{1}{c}{2-5}   & \multicolumn{1}{c}{1/3/9} \\
  \multicolumn{1}{c}{mildew}     & \multicolumn{1}{c}{35}   & \multicolumn{1}{c}{46}   & \multicolumn{1}{c}{3/3}  & \multicolumn{1}{c}{3-100} & \multicolumn{1}{c}{1/2/5} \\
  \multicolumn{1}{c}{alarm}      & \multicolumn{1}{c}{37}   & \multicolumn{1}{c}{46}   & \multicolumn{1}{c}{4/5}  & \multicolumn{1}{c}{2-4}   & \multicolumn{1}{c}{1/2/6} \\
  \multicolumn{1}{c}{hailfinder} & \multicolumn{1}{c}{56}   & \multicolumn{1}{c}{66}   & \multicolumn{1}{c}{4/16} & \multicolumn{1}{c}{2-11}  & \multicolumn{1}{c}{1/1.5/17} \\
  \multicolumn{1}{c}{munin1}     & \multicolumn{1}{c}{186} & \multicolumn{1}{c}{273} & \multicolumn{1}{c}{3/15} & \multicolumn{1}{c}{2-21}  & \multicolumn{1}{c}{1/3/15} \\
  \multicolumn{1}{c}{pigs}       & \multicolumn{1}{c}{441}  & \multicolumn{1}{c}{592}  & \multicolumn{1}{c}{2/39} & \multicolumn{1}{c}{3-3}   & \multicolumn{1}{c}{1/2/41} \\
  \multicolumn{1}{c}{link}       & \multicolumn{1}{c}{724}  & \multicolumn{1}{c}{1125} & \multicolumn{1}{c}{3/14} & \multicolumn{1}{c}{2-4}   & \multicolumn{1}{c}{0/2/17}  \\
  \hline
  \end{tabularx}
\end{table}

In this section, we conduct an experimental comparison of H2PC against
MMHC on several benchmarks with various data sizes. All the data sets
used for the empirical experiments are sampled from eight well-known
BNs that have been previously used as benchmarks for BN learning
algorithms (see Table \ref{tab:datasets} for details). We do not claim
that those data sets resemble real-world problems, however, they make
it possible to compare the outputs of the algorithms with the known
structure. All BN benchmarks (structure and probability tables) were
downloaded from the \textit{bnlearn} repository\footnote{
\emph{http://www.bnlearn.com/bnrepository}} \cite{Scutari10}. Six sample sizes have
been considered: 50, 100, 200, 500, 1500 and 5000. All experiments are
repeated 10 times for each sample size and each BN. We investigate the
behavior of both algorithms using the same parametric tests as a
reference. H2PC was implemented in \textit{R} \cite{R10} and integrated
into the \textit{bnlearn} \textit{R} package developed by
\cite{Scutari10}. The source code of H2PC as well as all data sets used
for the empirical tests are publicly available \footnote{
\emph{http://www710.univ-lyon1.fr/$\sim$aaussem/Software.html}}. The
threshold considered for the type I error of the test is 0.05. Our
experiments were carried out on PC with Intel(R) Core(TM) i5-2520M CPU
@2,50 GHz 4Go RAM running under Windows 7 32 bits.

\begin{figure}
\begin{center}$
  \begin{array}{cc}
  \includegraphics[clip, trim = 0.05in 0.1in 0.1in 0.1in]{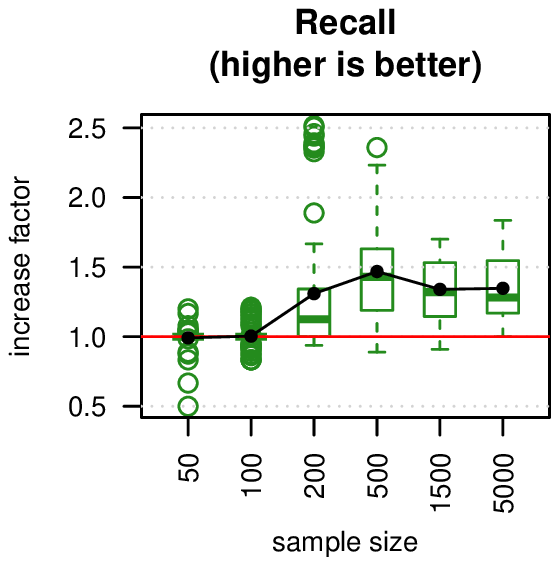} &
  \includegraphics[clip, trim = 0.05in 0.1in 0.1in 0.1in]{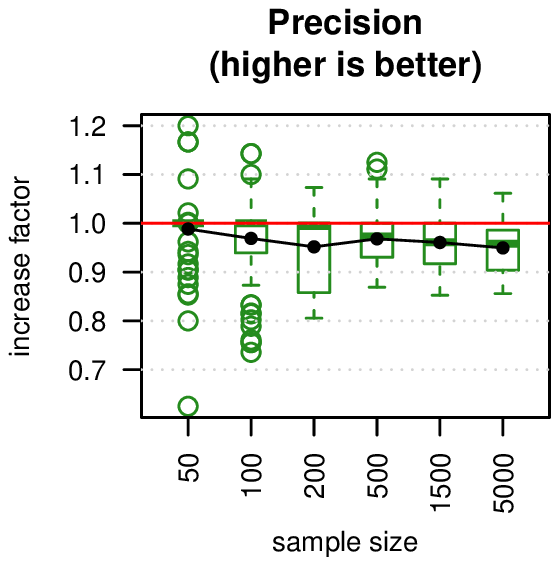} \\
  \includegraphics[clip, trim = 0.05in 0.1in 0.1in 0.1in]{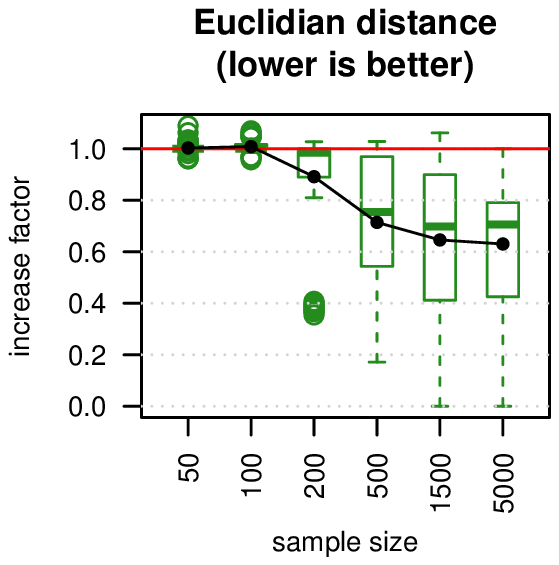} &
  \includegraphics[clip, trim = 0.05in 0.1in 0.1in 0.1in]{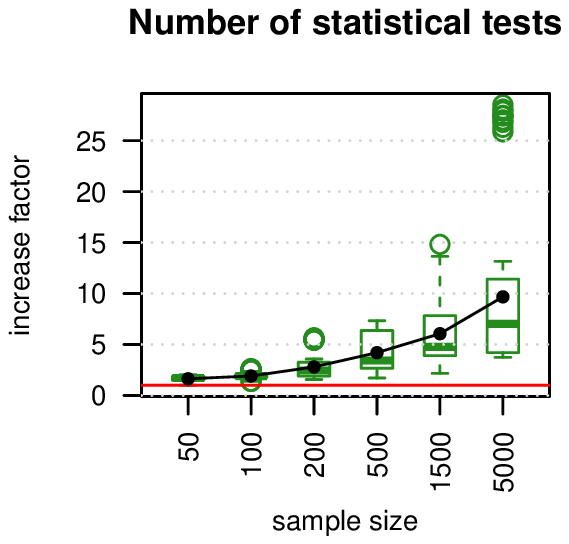} \\
   \includegraphics[clip, trim = 0.05in 0.1in 0.1in 0.1in]{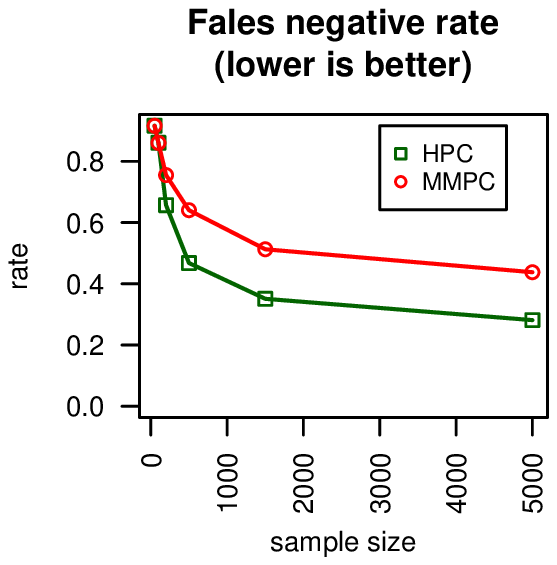} &
  \includegraphics[clip, trim = 0.05in 0.1in 0.1in 0.1in]{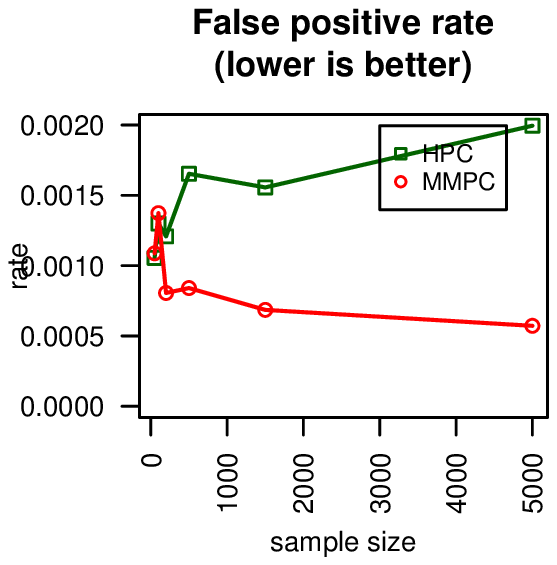} \\
  \end{array}$
  \end{center}
  \caption{Quality of the skeleton obtained with HPC over that obtained with MMPC before the SS phase. Results are averaged over the 8 benchmarks.}
  \label{fig:h2pc-skeleton-scores}
\end{figure}

We first investigate the quality of the skeleton returned by H2PC
during the CB phase. To this end, we measure the false positive edge
ratio, the precision (i.e., the number of true positive edges in the
output divided by the number of edges in the output), the recall (i.e.,
the number of true positive edges divided the true number of edges) and
a combination of precision and recall defined as $\sqrt{(1-precision)^2
+ (1-recall)^2}$, to measure the Euclidean distance from perfect
precision and recall, as proposed in \cite{Pen07}. Second, to assess
the quality of the final DAG output at the end of the SS phase, we
report the five performance indicators \cite{Scutari12} described below:

\begin{itemize}
	\item the posterior density of the network for the data it was
  learned from, as a   measure of goodness of fit. It is known as the
  Bayesian Dirichlet equivalent score (BDeu) from \cite{Heckerman95,Bun91} and
  has a single parameter, the equivalent sample size, which can be
  thought of as the size of an imaginary sample supporting the prior
  distribution. The equivalent sample size was set to 10 as suggested
  in \cite{Koller09};
  \item the BIC score \cite{Schwarz78} of the network for the data it
  was learned from, again as a measure of goodness of fit;
  \item the posterior density of the network for a new data set, as a
  measure of how well the network generalizes to new data; 
  \item  the BIC score of the network for a new data set, again as a
  measure of how well the network generalizes to new data;
  \item the Structural Hamming Distance (SHD) between the learned and
  the true structure of the network, as a measure of the quality of the
  learned dependence structure. The SHD between two PDAGs is defined as
  the number of the following operators required to make the PDAGs
  match: add or delete an undirected edge, and add, remove, or reverse
  the orientation of an edge.
\end{itemize}

For each data set sampled from the true probability distribution of the benchmark, we
first learn a network structure with the H2PC and MMHC and then we compute the
relevant performance indicators for each pair of network structures. 
The data set used to assess how well the network generalizes to new data
is generated again from the true probability structure of the benchmark
networks and contains 5000 observations. 

Notice that using the BDeu score as a metric of
reconstruction quality has the following two problems. First, the score
corresponds to the a posteriori probability of a network only under
certain conditions (e.g., a Dirichlet distribution of the hyper
parameters); it is unknown to what degree these assumptions hold in
distributions encountered in practice. Second, the score is highly sensitive to the
equivalent sample size (set to 10 in our experiments) and depends on the network priors used. 
Since, typically, the same arbitrary value of this parameter is used both during learning and
for scoring the learned network, the metric favors algorithms that use
the BDeu score for learning. In fact, the BDeu score does not rely on
the structure of the original, gold standard network at all; instead it
employs several assumptions to score the networks. For those reasons,
in addition to the score we also report the BIC score and the SHD
metric. 

In Figure \ref{fig:h2pc-skeleton-scores}, we report the quality of the skeleton obtained with HPC over that obtained with MMPC (before the SS phase) as a function of the sample size. Results for each benchmark are not shown here in detail due to space
restrictions. For sake of conciseness, the performance values are averaged over the 8 benchmarks depicted in Table \ref{tab:datasets}.
The increase factor for a given performance indicator is expressed as the ratio of the performance value obtained with HPC over that obtained with MMPC (the gold standard). Note that for some indicators, an increase is actually not an improvement 
but is worse (e.g., false positive rate, Euclidean distance). For clarity, 
we mention explicitly on the subplots whether an increase factor $>1$
should be interpreted as an improvement or not. Regarding the quality of the superstructure, the
advantages of HPC against MMPC are noticeable. As observed, HPC consistently increases the recall and reduces the rate of 
false negative edges. As expected this benefit comes at a little expense in terms
of false positive edges. HPC also improves the Euclidean distance
from perfect precision and recall on all benchmarks, while increasing the
number of independence tests and thus the running time in the CB phase (see
number of statistical tests). It is worth noting that HPC is capable of
maintaining the mean false positive edge increase (with respect to
MMPC) under $2 \cdot 10^{-3}$  while reducing by 30\%
the Euclidean distance in the range 500-5000 samples. These results are very much in line with other experiments presented in
\cite{Morais10b,Villanueva12}.

\begin{figure}
  \begin{center}$
  \begin{array}{cc}
  \includegraphics[clip, trim = 0.05in 0.1in 0.1in 0.1in]{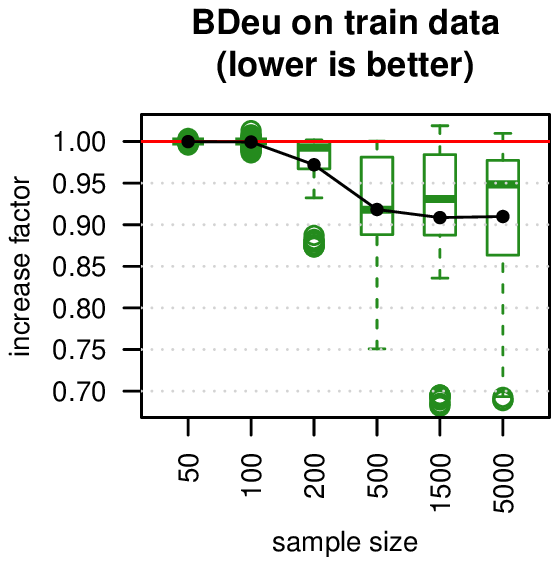} & 
  \includegraphics[clip, trim = 0.05in 0.1in 0.1in 0.1in]{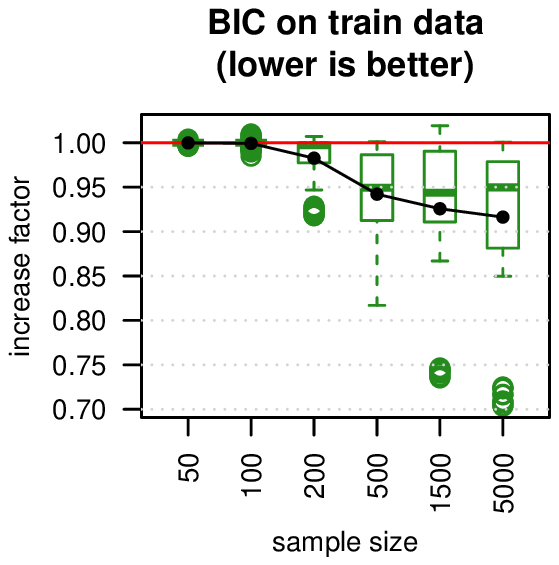}\\
  \includegraphics[clip, trim = 0.05in 0.1in 0.1in 0.1in]{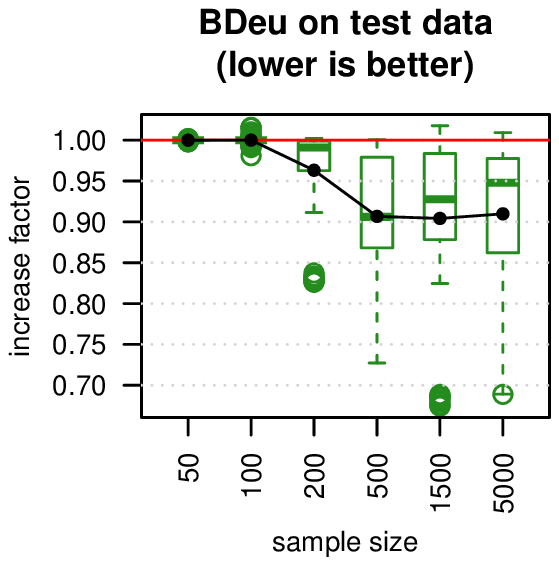} &
  \includegraphics[clip, trim = 0.05in 0.1in 0.1in 0.1in]{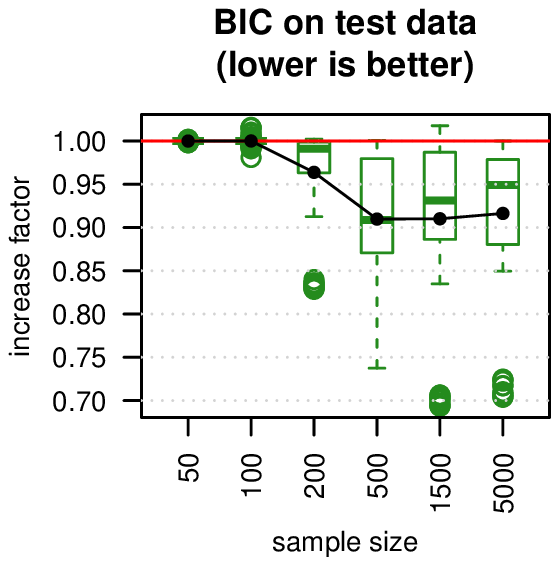}\\
  \includegraphics[clip, trim = 0.05in 0.1in 0.1in 0.1in]{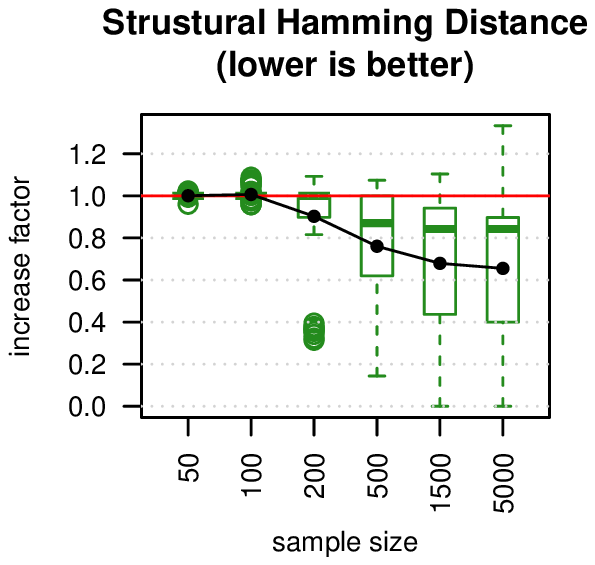} &
  \includegraphics[clip, trim = 0.05in 0.1in 0.1in 0.1in]{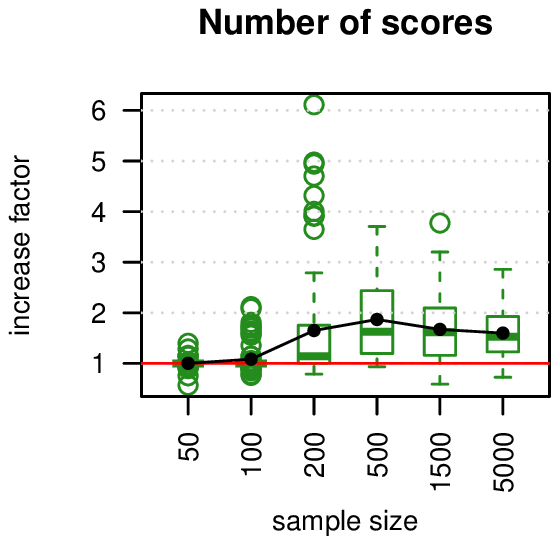} \\
  \end{array}$
  \end{center}
  \caption{Quality of the final DAG obtained with H2PC over that obtained with MMHC (after the SS phase). Results are averaged over the 8 benchmarks.}
  \label{fig:h2pc-scores}
\end{figure}

In Figure \ref{fig:h2pc-scores}, we report the quality of the final DAG obtained with H2PC over that obtained with MMHC (after the SS phase)
as a function of the sample size. Regarding BDeu and BIC on both training and test data,  
the improvements are noteworthy. The results in terms of
goodness of fit to training data and new data using H2PC clearly
dominate those obtained using MMHC, whatever the sample size considered,
hence its ability to generalize better. Regarding the quality of the
network structure itself (i.e., how close is the DAG  to the true
dependence structure of the data), this is pretty much a dead heat between the 2 algorithms on small sample sizes (i.e., 50 and 100), however 
we found H2PC to perform significantly better on larger
sample sizes. The SHD increase factor decays rapidly (lower is better) as the sample size increases. As far as the overall
running time performance is concerned, we see from Table \ref{fig:h2pc-scores}
that both methods have a tendency to work comparatively well for small sample sizes (i.e., less than 200). The total running time with H2PC
with 5000 samples is 8 times slower on average than that of MMHC. Overall, it appears that the running time increase factor grows somewhat linearly with the sample size.
Nonetheless,  it is worth mentioning that our implementation of MMHC in the \textit{bnlearn} package employs several heuristics to speed up learning that are not yet implemented in H2PC.
This leads to some loss of efficiency compared to MMHC due to redundant calculations. 
Notice that the optimization of the HPC code is currently being undertaken to allow for fair comparisons with MMHC.

Overall, H2PC compares favorably to MMHC. It has consistently lower
generalization error on all data sets. Large values of the recall do
not cause much rise in precision while maintaining the total running
time under control. This experiment indicates that MMPC should be best
suited in terms of performance when coupled to an optimal SS BN
learning method discussed in \cite{Perrier08,Kojima10}.

\begin{table}[h]
  \caption{Total running time increase factor (H2PC/MMHC).}
  \label{tab:total.time}
  \begin{tabularx}{\textwidth}{l X X X X X X X}
    \hline
    \multicolumn{1}{c}{\multirow{2}{*}{Network}} & \multicolumn{6}{c}{Sample Size} \\ \cline{2-7}
    \multicolumn{1}{c}{} & \multicolumn{1}{c}{50} & \multicolumn{1}{c}{100} & \multicolumn{1}{c}{200} & \multicolumn{1}{c}{500} & \multicolumn{1}{c}{1500} & \multicolumn{1}{c}{5000} \\
    \hline
    child       & 1.13 \(\pm\)0.1           & 1.28 \(\pm\)0.2 & 1.54 \(\pm\)0.2 & 2.38 \(\pm\)0.2 & 2.65 \(\pm\)0.2 & 3.08 \(\pm\)0.4 \\
    insurance   & 1.21 \(\pm\)0.2           & 1.35 \(\pm\)0.2 & 2.03 \(\pm\)0.1 & 3.83 \(\pm\)0.2 & 5.55 \(\pm\)0.4 & 7.38 \(\pm\)0.6 \\
    mildew      & 0.71 \(\pm\)0.2  					& 1.05 \(\pm\)0.2 & 1.10 \(\pm\)0.1 & 1.23 \(\pm\)0.1 & 1.63 \(\pm\)0.1 & 3.19 \(\pm\)0.3 \\
    alarm       & 1.17 \(\pm\)0.1           & 1.39 \(\pm\)0.1 & 1.86 \(\pm\)0.1 & 2.28 \(\pm\)0.1 & 2.93 \(\pm\)0.3 & 3.41 \(\pm\)0.4 \\
    hailfinder  & 1.09 \(\pm\)0.1           & 1.30 \(\pm\)0.1 & 1.61 \(\pm\)0.1 & 2.17 \(\pm\)0.1 & 2.82 \(\pm\)0.2 & 3.35 \(\pm\)0.3 \\
    munin1      & 1.09 \(\pm\)0.0           & 1.29 \(\pm\)0.1 & 1.36 \(\pm\)0.1 & 2.01 \(\pm\)0.1 & 4.28 \(\pm\)0.2 & 12.88 \(\pm\)0.7 \\
    pigs        & 1.41 \(\pm\)0.1           & 1.41 \(\pm\)0.1 & 4.65 \(\pm\)0.2 & 5.32 \(\pm\)0.2 & 6.51 \(\pm\)0.2 & 9.70 \(\pm\)0.3 \\
    link        & 1.57 \(\pm\)0.0           & 2.13 \(\pm\)0.1 & 2.86 \(\pm\)0.1 & 6.07 \(\pm\)0.2 & 11.07 \(\pm\)1.1 & 23.35 \(\pm\)0.9 \\
    \hline
    all         & 1.17 \(\pm\)0.3           & 1.40 \(\pm\)0.3 & 2.13 \(\pm\)1.1 & 3.16 \(\pm\)1.6 & 4.68 \(\pm\)2.9 & 8.29 \(\pm\)6.7 \\
    \hline
  \end{tabularx}
\end{table}

\section{Discussion}

Our prime conclusion is that H2PC is a promising approach to
constructing BN structures. The performances of HPC raises interesting
possibilities in the context of hybrid methods. It emphasizes that
concentrating on higher recall values while keeping the false positive
rate as low as possible  pays off in terms of goodness of fit and
structure accuracy. 

The focus of our study was on the efficiency of the heuristics the
learning algorithms are based on, i.e., the maximization algorithms
used in score-based algorithms combined with the techniques for
learning the dependence structure associated with each node in CB
algorithms. The influence of the other components of the overall
learning strategy, such as the conditional independence tests (and the
associated type I error threshold) or the network scores (and the
associated parameters, such as the equivalent sample size), was not
investigated. 

The conclusions of our studies should be applicable to
the shrinkage test that is more robust to small sample sizes, and to
the permutation mutual information test for large samples. Tsamardinos et al. 
\cite{Tsamardinos10} recently showed that the use of exact tests based
on (semi-parametric) permutation procedures lead to more robust
structural learning, while being only 10-20 times slower than the
asymptotic tests for small sample sizes. Similarly, Scutari \cite{Scutari11}
investigated the behavior of permutation conditional independence tests
and tests based the permutation Pearson's $\chi^2$ test, the
permutation mutual information test, and the shrinkage test based on
the estimator for the mutual information. Based on a single BN
benchmark, they showed that permutation tests result in better network
structures than the corresponding parametric tests in terms of goodness
of fit. However, the output graphs are often not as close to the true
network structure as the ones learned with the corresponding parametric
tests. Shrinkage tests, on the other hand, outperform both parametric
and permutation tests in the quality of the network structure itself,
which is closer to the true dependence structure but do not fit the
data as well as the networks learned with the corresponding maximum
likelihood tests. So, there is no clear picture as to which test should
be employed on a given data set. This is still an open question.

As noted by \cite{Perrier08}, it is possible to reduce the complexity
of an optimal search of an exponential factor by using a structural
constraint such as a super-structure on condition that this
super-structure is sound (i.e., includes the true graph as a subgraph).
Under this assumption, the accuracy of the resulting graph may greatly
improve according. Consequently, more attention should be paid to
learning sound superstructures rather than true skeleton from data as
both speed and accuracy should be expected with more sophisticated SS
search strategies than the greedy HC used in our study. Although sound
super-structures are easier to learn for high values of type I error,
such values produce denser structures with many extra edges, thereby
resulting in high computational overheads. Therefore, relaxing the type
I error of the tests is not a solution. \cite{Kojima10} show that a
small change in the type I error in MMPC yields a dramatical increase
of the computational burden involved in their hybrid procedure, with
almost no gain in accuracy. The key is to keep the false positive
rate small while controlling the false missing rate.

Finally, it is worth mentioning that neither MMPC nor HPC were
optimized in this work to learn global superstructures as both MMPC and
HPC are run independently on each node without keeping track of the
dependencies found previously. This leads to some loss of efficiency
due to redundant calculations. The reason is that they were initially
designed to infer a local network around a target node. An optimized
version of HPC for super-structure discovery was developed in
\cite{Villanueva12}. The optimizations were done in order to get a
global method and to lower the computational cost of HPC, while
maintaining its performance. These optimizations include the use of a
cache to store the (in)dependencies and the use of a global structure.
This optimizations reduce the computational cost of HPC by 30\% on
average according to the authors.

\section{Conclusion}

We discussed a hybrid algorithm for BN structure learning called Hybrid
HPC (H2PC). Our extensive experiments show that H2PC outperforms MMHC
in terms of goodness of fit to training data and new data as well,
hence its ability to generalize better, with little overhead in terms of
of running time over MMHC. The optimization of the HPC code is currently being undertaken. 
Regarding the quality of the
network structure itself (i.e., how close is the DAG to the true
dependence structure of the data), we found H2PC to outperform MMHC by a significant margin. 
More importantly, our experimental results show a clear benefit in terms of
edge recall without sacrificing the number of extra edges, which is
crucial for the soundness of the super-structure used during the second
stage of hybrid methods like the ones proposed in \cite{Perrier08,Kojima10}.
Though not discussed here, a topic of considerable interest would be to
ascertain which independence test is most suited to the data at hand.
This needs further substantiation through more experiments and analysis.

\section*{Acknowledgments}

The authors thank Marco Scutari for sharing his \textit{bnlearn}
package in \textit{R}. The experiments reported here were performed on
computers funded by a French Institute for Complex Systems (IXXI) grant. 

\begin{small}
\bibliography{biblio}

\begin{thebibliography}{10}

\bibitem{Agresti02}
A.~Agresti.
\newblock {\em Categorical Data Analysis}.
\newblock Wiley, 2nd edition, 2002.

\bibitem{Aliferis10a}
C.~F. Aliferis, A.~R. Statnikov, I.~Tsamardinos, S.~Mani, and X.~D. Koutsoukos.
\newblock Local causal and markov blanket induction for causal discovery and
  feature selection for classification part i: Algorithms and empirical
  evaluation.
\newblock {\em Journal of Machine Learning Research}, 11:171--234, 2010.

\bibitem{ArmenT11}
A.~P. Armen and I.~Tsamardinos.
\newblock A unified approach to estimation and control of the false discovery
  rate in bayesian network skeleton identification.
\newblock In {\em European Symposium on Artificial Neural Networks, ESANN'11},
  2011.

\bibitem{Aussem12a}
A.~Aussem, S.~{Rodrigues de Morais}, and M.~Corbex.
\newblock Analysis of nasopharyngeal carcinoma risk factors with bayesian
  networks.
\newblock {\em Artificial Intelligence in Medicine}, 54(1), 2012.

\bibitem{Aussem10c}
A.~Aussem, A.~Tchernof, S.~{Rodrigues de Morais}, and S.~Rome.
\newblock Analysis of lifestyle and metabolic predictors of visceral obesity
  with bayesian networks.
\newblock {\em BMC Bioinformatics}, 11:487, 2010.

\bibitem{Brown08}
L.~E. Brown and I.~Tsamardinos.
\newblock A strategy for making predictions under manipulation.
\newblock {\em JMLR: Workshop and Conference Proceedings}, 3:35--52, 2008.

\bibitem{Bun91}
W.~Buntine.
\newblock Theory refinement on {B}ayesian networks.
\newblock In Bruce {D'Ambrosio}, Philippe Smets, and Piero Bonissone, editors,
  {\em Proceedings of the 7th Conference on Uncertainty in Artificial
  Intelligence}, pages 52--60, San Mateo, CA, USA, July 1991. Morgan Kaufmann
  Publishers.

\bibitem{Cawley08}
Gavin Cawley.
\newblock Causal and non-causal feature selection for ridge regression.
\newblock {\em JMLR: Workshop and Conference Proceedings}, 3, 2008.

\bibitem{Cheng02}
Jie Cheng, Russell Greiner, Jonathan Kelly, David~A. Bell, and Weiru Liu.
\newblock Learning {B}ayesian networks from data: An information-theory based
  approach.
\newblock {\em Artif. Intell.}, 137(1-2):43--90, 2002.

\bibitem{Chickering02}
David~Maxwell Chickering.
\newblock Optimal structure identification with greedy search.
\newblock {\em Journal of Machine Learning Research}, 3:507--554, 2002.

\bibitem{Byron08}
Byron Ellis and Wing~Hung Wong.
\newblock Learning causal bayesian network structures from experimental data.
\newblock {\em Journal of the American Statistical Association}, 103:778--789,
  2008.

\bibitem{Friedman99}
N.L Friedman, I.~Nachman, and D.~Pe'er.
\newblock Learning bayesian network structure from massive datasets: the
  "sparse candidate" algorithm.
\newblock In Kathryn~B. Laskey and Henri Prade, editors, {\em Proceedings of
  the 15th Conference on Uncertainty in Artificial Intelligence}, pages 21--30.
  Morgan Kaufmann Publishers, 1999.

\bibitem{Heckerman95}
D~Heckerman, D~Geiger, and D.M Chickering.
\newblock Learning bayesian networks: The combination of knowledge and
  statistical data.
\newblock {\em Machine Learning}, {20}({3}):{197--243}, {1995}.

\bibitem{Koivisto04}
M.~Koivisto and K.~Sood.
\newblock Exact bayesian structure discovery in bayesian networks.
\newblock {\em Journal of Machine Learning Research}, 5:549--573, 2004.

\bibitem{Kojima10}
K.~Kojima, E.~Perrier, S.~Imoto, and S.~Miyano.
\newblock Optimal search on clustered structural constraint for learning
  bayesian network structure.
\newblock {\em Journal of Machine Learning Research}, 11:285--310, 2010.

\bibitem{Koller09}
D.~Koller and N.~Friedman.
\newblock {\em Probabilistic Graphical Models: Principles and Techniques}.
\newblock MIT Press, 2009.

\bibitem{Moore03}
Andrew Moore and Weng-Keen Wong.
\newblock Optimal reinsertion: A new search operator for accelerated and more
  accurate {B}ayesian network structure learning.
\newblock In T.~Fawcett and N.~Mishra, editors, {\em Proceedings of the 20th
  International Conference on Machine Learning (ICML '03)}, August 2003.

\bibitem{Pen07}
J.M. {Pe{\~n}a}, R.~Nilsson, J.~Bj{\"o}rkegren, and J.~Tegn{\'e}r.
\newblock Towards scalable and data efficient learning of {M}arkov boundaries.
\newblock {\em International Journal of Approximate Reasoning}, 45(2):211--232,
  2007.

\bibitem{Pea88}
J.~Pearl.
\newblock {\em Probabilistic Reasoning in Intelligent Systems: Networks of
  Plausible Inference.}
\newblock Morgan Kaufmann, San Francisco, CA, USA, 1988.

\bibitem{Pena08}
J.~Pe{\~n}a.
\newblock Learning gaussian graphical models of gene networks with false
  discovery rate control.
\newblock In {\em Proceedings of 6th European Conference on Evolutionary
  Computation, Machine Learning and Data Mining in Bioinformatics}, pages
  165--176, 2008.

\bibitem{Pen12}
J.~Pe{\~n}a.
\newblock Finding consensus bayesian network structures.
\newblock {\em Journal of Artificial Intelligence Research}, 42:661--687, 2012.

\bibitem{Perrier08}
E.~Perrier, S.~Imoto, and S.~Miyano.
\newblock Finding optimal bayesian network given a super-structure.
\newblock {\em Journal of Machine Learning Research}, 9:2251--2286, 2008.

\bibitem{Morais10b}
S.~{Rodrigues de Morais} and A.~Aussem.
\newblock An efficient learning algorithm for local bayesian network structure
  discovery.
\newblock In {\em European Conference on Machine Learning and Principles and
  Practice of Knowledge Discovery in Databases, ECML-PKDD'10}, pages 164--169,
  2010.

\bibitem{Morais10a}
S.~{Rodrigues de Morais} and A.~Aussem.
\newblock A novel {M}arkov boundary based feature subset selection algorithm.
\newblock {\em Neurocomputing}, 73:578--584, 2010.

\bibitem{Schwarz78}
G.~E. Schwarz.
\newblock Estimating the dimension of a model.
\newblock {\em Journal of Biomedical Informatics}, 6(2):461--464, 1978.

\bibitem{Scutari10}
M.~Scutari.
\newblock Learning bayesian networks with the bnlearn {R} package.
\newblock {\em Journal of Statistical Software}, 35(3):1--22, 2010.

\bibitem{Scutari12}
M.~Scutari and A.~Brogini.
\newblock Bayesian network structure learning with permutation tests.
\newblock {\em To appear in Communications in Statistics - Theory and Methods},
  2012.

\bibitem{Scutari11}
Marco Scutari.
\newblock {\em Measures of Variability for Graphical Models}.
\newblock PhD thesis, School in Statistical Sciences, University of Padova,
  2011.

\bibitem{Silander06}
T.~Silander and P.~Myllymaki.
\newblock Simple approach for finding the globally optimal {B}ayesian network
  structure.
\newblock In {\em Proceedings of the 22nd Conference on Uncertainty in
  Artificial Intelligence ({UAI}-06)}, pages 445--452, 2006.

\bibitem{Spi00}
P.~Spirtes, C.~Glymour, and R.~Scheines.
\newblock {\em Causation, Prediction, and Search}.
\newblock The {MIT} Press, 2nd edition, 2000.

\bibitem{R10}
R~Development~Core Team.
\newblock R: A language and environment for statistical computing.
\newblock {\em R Foundation for Statistical Computing, Vienna, Austria}, 2010.

\bibitem{Tsamardinos03}
I.~Tsamardinos, C.F. Aliferis, and A.R. Statnikov.
\newblock Algorithms for large scale {M}arkov blanket discovery.
\newblock In {\em Florida Artificial Intelligence Research Society Conference
  FLAIRS'03}, pages 376--381, 2003.

\bibitem{Tsamardinos10}
I.~Tsamardinos and G.~Borboudakis.
\newblock Permutation testing improves bayesian network learning.
\newblock In {\em Machine Learning and Knowledge Discovery in Databases,
  European Conference, ECML PKDD 2010}, pages 322--337, 2010.

\bibitem{Tsamardinos08}
I.~Tsamardinos and L.~E. Brown.
\newblock Bounding the false discovery rate in local {B}ayesian network
  learning.
\newblock In {\em Proceedings AAAI National Conference on AI AAAI'08}, pages
  1100--1105, 2008.

\bibitem{Tsamardinos06}
I.~Tsamardinos, L.E. Brown, and C.F. Aliferis.
\newblock The max-min hill-climbing {B}ayesian network structure learning
  algorithm.
\newblock {\em Machine Learning}, 65(1):31--78, 2006.

\bibitem{Villanueva12}
E.~Villanueva and C.D. Maciel.
\newblock Optimized algorithm for learning bayesian network superstructures.
\newblock In {\em Proceedings of the 2012 International Conference on Pattern
  Recognition Applications and Methods, ICPRAM'12}, 2012.

\end{thebibliography}
\end{small}

\end{document}